%% file: main.tex
\documentclass[conference]{IEEEtran}
\IEEEoverridecommandlockouts
\usepackage{cite}
\usepackage{amsmath,amssymb,amsfonts}
\usepackage{algorithm,algorithmicx}
\usepackage{algpseudocode}
\usepackage{graphicx}
\usepackage{textcomp}
\usepackage{rotating}
\usepackage{multirow}

\usepackage{hyperref}

\usepackage{url} 
\usepackage{xcolor}
\def\BibTeX{{\rm B\kern-.05em{\sc i\kern-.025em b}\kern-.08em
    T\kern-.1667em\lower.7ex\hbox{E}\kern-.125emX}}
\begin{document}

\newcommand{\asaf}[1]{{\textcolor{red}{[Asaf: #1]}}}
\newcommand{\guy}[1]{{\textcolor{orange}{[Guy: #1]}}}
\newcommand{\mosh}[1]{{\textcolor{teal}{[Moshe: #1]}}}
\newcommand{\MethodName}{FOOD }
\newcommand{\argmax}{\mathop{\mathrm{arg\,max}}}

\title{FOOD: Fast Out-Of-Distribution Detector}

\input{Authors}

\maketitle

\input{Abstract}
\begin{IEEEkeywords}
Neural network, Out-of-Distribution, Representations
\end{IEEEkeywords}

\input{Intro}
\input{Background}
\input{Related_Work}
\input{Method}

\input{Exp_setup}
\input{Results}
\input{Conclusions}

\bibliography{Bibliography.bib}
\bibliographystyle{IEEEtran}

\end{document}

%% file: Authors.tex
\author{\IEEEauthorblockN{Guy Amit*, Moshe Levy*,  Ishai Rosenberg, Asaf Shabtai, and Yuval Elovici}
    \IEEEauthorblockA{\textit{Department of Software and Information Systems Engineering} \\
    \textit{Ben-Gurion University of the Negev}\\
    Beer-Sheva, Israel \\
    \{guy5, moshe5, ishairos\}@post.bgu.ac.il, \{shabtaia, elovici\}@bgu.ac.il }
}

\maketitle

%% file: Abstract.tex
Deep neural networks (DNNs) perform well at classifying inputs associated with the classes they have been trained on, which are known as in-distribution inputs. 
However, out-of-distribution (OOD) inputs pose a great challenge to DNNs and consequently represent a major risk when DNNs are implemented in safety-critical systems. 
Extensive research has been performed in the domain of OOD detection. 
However, current state-of-the-art methods for OOD detection suffer from at least one of the following limitations: (1) increased inference time - this limits existing methods' applicability to many real-world applications, and (2) the need for OOD training data - such data can be difficult to acquire and may not be representative enough, thus limiting the ability of the OOD detector to generalize.
In this paper, we propose \MethodName -- Fast Out-Of-Distribution detector -- an extended DNN classifier capable of efficiently detecting OOD samples with minimal inference time overhead. 
Our architecture features a DNN with a final Gaussian layer combined with the log likelihood ratio statistical test and an additional output neuron for OOD detection.
Instead of using real OOD data, we use a novel method to craft artificial OOD samples from in-distribution data, which are used to train our OOD detector neuron.
We evaluate FOOD's detection performance on the SVHN, CIFAR-10, and CIFAR-100 datasets.
Our results demonstrate that in addition to achieving state-of-the-art performance, \MethodName is fast and applicable to real-world applications.\footnote{Our code is available online: \url{https://github.com/guyAmit/GLOD}}

\let\thefootnote\relax\footnotetext{*Equal contribution}

%% file: Intro.tex
\section{\label{sec:intro}Introduction}
Deep neural networks (DNNs) have been successfully used in many classification tasks in various domains, including computer vision~\cite{xu2015show}, autonomous driving~\cite{Bojarski2016EndTE} and speech recognition~\cite{lecun2015deep}.
While capable of learning complex patterns, DNNs are limited in their ability to correctly classify samples that are not associated with any of the training data classes. 
Such samples are known as out-of-distribution (OOD) samples, and they have been identified as a major risk in AI~\cite{amodei2016concrete,mohseni2019practical}. 

The capability of detecting that a sample is OOD is even more significant in safety-critical systems, such as autonomous driving systems.
What makes autonomous driving systems particularly challenging for OOD detection is that significant operational overhead cannot be tolerated, and the consequences of an error are severe~\cite{mohseni2019practical}.
Previous studies addressed the OOD challenge with various real-world design considerations.
Some studies suggested OOD detection models that perform fast inference~\cite{mohseni2020self,hendrycks2018deep}, however the proposed detection models depend on access to OOD samples for training, which may not be possible in real-world applications.
In addition, the OOD samples used for training may not be representative, and thus the OOD detector cannot generalize well on unexpected inputs~\cite{shafaei2019less}. 
There are methods that refrain from using OOD samples to train their models~\cite{hsu2020generalized,sastry2019detecting}, however those models suffer from increased inference time.

To both achieve fast inference and eliminate the need to acquire OOD samples for training, we designed our method guided by two assumptions: (1) The inherent OOD detection capability in a DNN is enhanced when it is trained to model the in-distribution classes as density functions. (2) The density functions can be exploited to simulate the behavior of the DNN when encountering OOD samples. 

We propose \MethodName --  Fast Out-of-Distribution Detector,
a DNN model featuring a final Gaussian~\cite{wan2018rethinking} layer that models a density function for each class. 
Using a log likelihood ratio statistical test on the outputs of the Gaussian layer, we generate samples that are perceived as OOD by the DNN. 
We employ those samples to train our OOD detector, which is implemented as an additional output neuron.

\textbf{Contributions.} We introduce FOOD, a fast OOD detector that does not require OOD samples for training or hyperparameter tuning.
We perform an inference time comparison, showing that current state-of-the-art methods suffer from a trade-off between inference time overhead and OOD detection performance. 
We evaluate \MethodName on commonly used benchmark datasets for OOD detection: SVHN~\cite{netzer2011reading}, CIFAR-10~\cite{krizhevsky2009learning},
and CIFAR-100~\cite{krizhevsky2009learning}. Our evaluation demonstrates FOOD's state-of-the-art results, and our inference time comparison indicates that these results do not come at the cost of inference time overhead, which is minimal.
Our contributions can be summed up as follows:
\begin{enumerate}
    \item A fast OOD detection method that achieves state-of-the-art performance without the need for OOD data for training.
    \item A novel method for generating samples that are perceived by the DNN as OOD; this data generation process is insensitive to manual parameter choices and does not require adjustments when used on different datasets.  
\end{enumerate}

%% file: Background.tex
\section{\label{sec:background}Background}

In this section, we provide the following: (1) a definition of the OOD problem and our requirements for a solution, and (2) relevant background on Gaussian distribution use in DNN architectures.

\subsection{Problem Formalization} 
In a DNN classifier without OOD detection capability, there is an underlying assumption that a test sample $x$ is associated with one of the $C$ classes in the training dataset. By denoting the class that $x$ is associated with as $c(x)$, the classification can be expressed as: 
\begin{equation} \argmax_{i\in\{1,\ldots,C\}} p(i|x  \wedge \exists j ; c(x) = j)
\label{eq1}
\end{equation}
OOD detection necessitates the option of classifying the input $x$ as none of the learned classes associated with the training dataset.
In this case, the classification can be expressed as:
\begin{equation}
 \argmax_{i\in\{1,\ldots,C\}\cup\{OOD\}} p(i| x)
\label{eq2}
\end{equation}
\\
We propose a novel OOD detection method that meets the following requirements:

\begin{enumerate}
    \label{restrictions}
    \item \textit{Fast inference time - } The method must be able to be applied in scenarios where inference time is limited. Therefore, no significant inference time overhead is permitted.
    \item \textit{In-distribution data only - } The method must only use in-distribution data for training. 
\end{enumerate}

\subsection{Gaussian Distribution in DNN}
The multivariate Gaussian function is integrated in DNN methods and architectures in numerous ways~\cite{van2014factoring, wang2017maximum, wan2018rethinking}. For computational reasons, usually the log values of the likelihoods are calculated instead of the likelihoods' values themselves.
Given the parameters $\Sigma\in \mathcal{R}^{d\times d}$ and $\mu \in \mathcal{R}^{d}$, it is defined as in~\cite[chap.~3]{anderson1958introduction}:
\begin{multline}
     \log\big(\mathcal{N}(x; \mu,\Sigma)  \big) = \\
    = -\frac{d}{2}\log(2\pi) -\frac{1}{2}\log(|\Sigma|) -\frac{1}{2}\big\Vert x-\mu \big\Vert^{2}_{{\Sigma}^{-1}}
    \label{eq3}
\end{multline}
\\
A Gaussian layer was proposed to improve DNN classification accuracy~\cite{wan2018rethinking} by employing the multivariate Gaussian distribution.
Each of the Gaussian layer's outputs corresponds to a density function with respect to the layer's input.
The Gaussian layer assumes a multivariate Gaussian distribution on its input, and each of its outputs match the log likelihood, as presented in Equation (\ref{eq3}).
In practice, it is based on a DNN architecture in which the classification layer (last fully connected layer) is replaced with a Guassian layer designed for classification.
The Gaussian layer contains the trainable parameters that define the distribution of each class.
For each class $c$, the layer assigns a score that corresponds to the log likelihood of the penultimate layer's output, modeled as a Gaussian:
\begin{equation}
         log(p(f(x)| c)) = \log\big(\mathcal{N}(f(x); \mu_{c},\Sigma_{c})  \big) 
    \label{eq4}
\end{equation}
where $\mu_{c}$ and $\Sigma_{c}$ are the layer's parameters for the $c^{th}$ class, and $f(x)$ is the penultimate layer's output.
Diagonal covariance matrices are used instead of full covariance matrices, resulting in significantly reduced computational complexity.
To train the parameters of the modified DNN, the authors used softmax paired with cross-entropy loss ($\mathcal{L}_{ce}$) on the outputs of the Gaussian layer.
In addition to $\mathcal{L}_{ce}$, a regularization term that maximizes the likelihood of the correct output is employed.
Given a sample $\{x_{i},y_{i}\}_{i=1}^{m}$(where $m$ is the sample size), the regularization term is defined as: 
\begin{multline}
\mathcal{R}_{ML}=-\log(p(\{f(x_{i}), y_{i}\}_{i=1}^m 
;\{\Sigma_{c}, \mu_{c}\}_{c=1}^{C})) = \\  = -\frac{1}{m}\sum_{c=1}^{C}\sum_{i=1}^{m}\mathrm{1}_{c=y_{i}}\cdot \log(\mathcal{N}(f(x);\mu_{c}, \Sigma_{c}))
\label{eq5}
\end{multline}
The model optimization target $\mathcal{L}$ can be summarized as: 
\begin{equation}
    \mathcal{L} = \mathcal{L}_{ce}+\lambda\mathcal{R}_{ML}
    \label{eq6}
\end{equation}
where $\lambda$ is a hyperparameter chosen using a validation set.

%% file: Related_Work.tex
\section{\label{sec:relatedworks}Related Work}
In this section, we (1) present methods from the field of OOD detection, (2) discuss how the methods meet the requirements presented in Section \ref{restrictions}, and (3) discuss the use of adversarial samples in OOD detection.
\subsection{OOD Detection Methods for Comparison}
\textbf{Max Softmax Probability (MSP)}~\cite{DBLP:conf/iclr/HendrycksG17}
\label{ref:MSP} - A baseline for OOD detection. It employs the predicted maximum class probability as a score to separate OOD samples from in-distribution samples.

\textbf{Mahalanobis Distance (MD)}~\cite{mahalanobis} - Based on analysis of DNN hidden representations and input preprocessing. 
This method operates under the assumption that in each layer's representation, the samples of each class are distributed as a Gaussian.
Each Gaussian is parameterized by a center vector and a shared full covariance (all classes share one covariance matrix).
The score for OOD detection is a weighted sum over the Mahalanobis distance calculated in each layer, whose weights are learned using OOD samples.
The authors of \cite{hsu2020generalized} proposed a modification of this method, in which input preprocessing is not performed, and a simple sum is used instead of a weighted sum; we denote this method as MD*.

\textbf{Outlier Exposer (OE)}~\cite{hendrycks2018deep} - A detection method in which the DNN is fine-tuned using OOD samples that are not from the OOD test dataset.
OE optimizes a loss function which includes an additional term, forcing the DNN to output equal predictions for all classes when encountering an OOD sample.
On inference, the MSP score is used for OOD detection.

\textbf{Generalized Odin (G.ODIN)}~\cite{hsu2020generalized} - A detection method built upon the Odin method \cite{liang2018enhancing}, eliminating its need for OOD samples.
Its architecture is comprised of temperature scaling and modification of the softmax function.
During inference, G.ODIN uses input preprocessing to craft an adversarial sample from the input.
The predicted maximum class probability of the output is used as the score for OOD detection.

\textbf{Self-Supervised Learning for OOD detection (SSL)}~\cite{mohseni2020self} - A detection method in which the last layer of the DNN is extended with several additional neurons that are trained using OOD samples (not from the OOD test dataset). 
The additional neurons are trained using an proprietary loss function, and their sum is used as a score for OOD detection.

\textbf{Gram matrices (GM)}~\cite{sastry2019detecting} - A detection method in which the correlation between different features in each layer's representation is extensively analyzed. 
Given a representation $F_{l}$ for layer $l$ and a maximal order $P$, the Gram matrix measure includes the calculation of the higher order Gram matrices: $(F_l^p F_{l}^{p^{T}})^{\frac{1}{p}}$,
for $p\in\{1,\ldots, P\}$,
where the power and root are element-wise operations.  
The deviation from the Gram matrix measure of the training data is used for OOD detection.

\textbf{MALCOM}~\cite{chen2020robust} - A detection method that maps the internal representations to the string domain. In the string domain,  a file compression technique is used to extract uncorrelated features from the DNN, which then serve as input to the most basic version of the MD method~\cite{mahalanobis}.

\subsection{Requirement Compliance Comparison}
\label{Considerations_Comparison}
\input{tables/table1}
In Table~\ref{table1:comperison}, we present a comparison of the various OOD methods with respect to our requirements, as specified in Section ~\ref{restrictions}.

MSP~\cite{DBLP:conf/iclr/HendrycksG17} only relies on the DNN's original output which was trained without any OOD data, so it meets the \textit{In-distribution data only} requirement. The output is used directly, and therefore it complies with the \textit{Fast inference time} requirement as well. 
MD~\cite{mahalanobis} and G.ODIN~\cite{hsu2020generalized} employ preprocessing to craft a perturbation of the input, which incurs significant inference time overhead, so they do not meet the \textit{Fast inference time} requirement.
MD also employs OOD samples to tune the OOD detection model, so it fails to comply with \textit{In-distribution data only}. 
However, MD* does not require OOD samples or perform input preprocessing, meaning that this modified version of MD meets both of our requirements. 
The SSL~\cite{mohseni2020self} and OE~\cite{hendrycks2018deep} methods are trained on OOD samples (not from the OOD test data), meaning that they fail to meet the \textit{In-distribution data only} requirement.
The MALCOM~\cite{chen2020robust} detection method is based on file compression techniques, which appear complicated but can be implemented efficiently, and since it does not rely on OOD samples, it complies with both of our requirements.
The GM~\cite{sastry2019detecting} method uses complex postprocessing calculations after the inference process has been performed.
This postprocess increases inference time, as shown in the Results section (Figure~\ref{fig:Speed_comparison}); therefore, it does not comply with the  \textit{Fast inference time} requirement.

\subsection{Adversarial Methods in OOD Detection}
Adversarial samples are in-distribution samples that have been carefully perturbed to shift their DNN prediction from the correct class to an incorrect one~\cite{szagdi}. There are different methods of crafting adversarial samples~\cite{Goodfellow2015ExplainingAH, carlini2017towards, akhtar2018threat}, each of which requires a different set of hyperparameters. 
Those hyperparameters guide the way in which the perturbation is crafted.
The same techniques can be used to achieve the opposite effect (i.e., to strengthen the confidence of the DNN's prediction in the correct class).
In general, it is possible to perturb the sample towards any direction of a differentiable score calculated by the DNN.
There are two main approaches in which OOD detection methods take advantage of adversarial methods:

In the first approach, it is assumed that adversarial perturbations have a different effect on OOD samples than on in-distribution samples.
Based on this assumption, during inference, some OOD detection methods craft an adversarial sample out of the input sample in order to enhance OOD detection~\cite{mahalanobis, liang2018enhancing, hsu2020generalized}. Each method crafts the perturbation according to its OOD detection score. 
In general, given a score $\mathcal{D}(\cdot)$, an input $x$, and a perturbation size $\epsilon$, the perturbed input $x_{per}$ is calculated as follows:
\begin{equation}
 x_{per} = x \pm \epsilon\cdot\frac{\partial \mathcal{D}(x)}{\partial x}
\end{equation}
The perturbation is either added or subtracted from the original $x$. Finally, the value of $\mathcal{D}(x_{per})$ is used for OOD detection.

Another approach for using adversarial samples in OOD detection is to replace the OOD samples used in the training phase.
In the paper proposing MD~\cite{mahalanobis},  the authors presented a version of MD that uses adversarial samples instead of OOD samples during the training phase. 
However, this MD version obtained suboptimal OOD detection results. 

%% file: tables/table1.tex

\begin{table}[b]
\centering
\caption{Requirement compliance}
\scalebox{0.98}{
\centering
\begin{tabular}{l|cccccccc|l}
                                                                    & {\begin{turn}{270}MSP\end{turn}}          & {\begin{turn}{270}MD\end{turn}}       & \multicolumn{1}{l}{\begin{turn}{270} 
MD*
\end{turn}} & \multicolumn{1}{l}{{\begin{turn}{270}OE\end{turn}}} & \multicolumn{1}{l}{{\begin{turn}{270}G.ODIN\end{turn}}} & \multicolumn{1}{l}{{\begin{turn}{270}SSL\end{turn}}} & \multicolumn{1}{l}{{\begin{turn}{270}GM\end{turn}}} & \multicolumn{1}{l}{{\begin{turn}{270}MALCOM\text{~}\end{turn}}} & {\begin{turn}{270}\textbf{FOOD}\end{turn}}         \\ \hline
\begin{tabular}[c]{@{}l@{}}In-Distribution\\ Data Only\end{tabular} & $\checkmark$ & $\times$ & $\checkmark$            & $\times$               & $\checkmark$               & $\times$                & $\checkmark$           & $\checkmark$               & $\checkmark$ \\ \hline
Fast Inference                                                      & $\checkmark$ & $\times$ & $\checkmark$            & $\checkmark$           & $\times$                   & $\checkmark$            & $\times$               & $ \checkmark$               & $\checkmark$
\end{tabular}
}

\label{table1:comperison}
\bigskip
\medskip
Table~\ref{table1:comperison}. Summary of related work,indicating each method's compliance to the specified requirements presented in Section ~\ref{Considerations_Comparison}. 
\end{table}

%% file: Method.tex
\section{Approach}
In this section, we introduce \MethodName - Fast OOD Detector. We start by presenting the high-level steps of the method's training, and then we describe each of the components of our method in detail.
The high-level steps for training \MethodName are: (1) adjust the DNN for OOD detection by replacing its final layer with a Gaussian~\cite{wan2018rethinking} based one, (2) generate adversarial samples which serve as artificial OOD samples, (3) extract the internal representations of the artificial OOD samples and the in-distribution training samples from the DNN,
(4) use the internal representations to train a logistic regression classifier that acts as an OOD detector; when the OOD detector is trained, the training samples are labeled as in-distribution, and the artificial OOD samples are labeled as OOD.
\subsection{Architecture \& Training}
\label{architecture}
Because of the \textit{Fast inference time} requirement, we developed a detection method that can be implemented efficiently as part of the DNN, without additional processing.
In the typical DNN classifier architecture, the final layer usually consists of linear transformation (a fully connected layer) and a softmax activation.
The softmax output can technically be employed as a density function, 
however studies have shown that the OOD detection capabilities of softmax are limited, and more sophisticated methods can surpass its OOD detection performance~ \cite{liang2018enhancing, DBLP:conf/iclr/HendrycksG17}.

Instead of using the softmax output, in this research we focus on the internal representations of the DNN, particularly the penultimate representation.
In prior research, representation analyses were performed to explain the relations in the geometric nature of DNN's internal representations~\cite{katzir2019detecting, mahalanobis}.
Those analyses showed that the samples of each class are naturally grouped in the internal representations, creating a clustering effect.
The deeper the layer is, the more obvious the clustering effect becomes, resulting in a complete separation between the classes in the penultimate representation.
This finding indicates that the penultimate layer has the potential to serve as the input to a density function.
We developed our architecture based on a method that employs a Gaussian layer as its final layer, instead of a fully connected one~\cite{wan2018rethinking}.
The Gaussian layer models each class ($c$) as a multivariate Gaussian, with its own trainable parameters - a positive semi-definite covariance matrix ($\Sigma_{c}\in \mathbb{R}^{d \times d}$) and a center vector ($\mu_{c}\in \mathbb{R}^{d}$). 
Since the likelihood term in the Gaussian layer (Equation \ref{eq3}) involves the calculation of  $\log(|\Sigma_{c}|)$ and ${\Sigma^{-1}_{c}}$, diagonal covariance matrices are used to improve numerical stability. 
Moreover, it allows efficient paralleled computation of the Gaussian's log likelihood.
Note that in our implementation, the covariance matrix is learned during training, in contrast to some other methods that employ Gaussians~\cite{yaseen2020preventing}. 

For simplicity, and unlike the procedure proposed in the original paper that introduced the Gaussian layer~\cite{wan2018rethinking}, we do not train our DNN from scratch, but rather fine-tune an existing trained DNN model.
We replace its last layer with a Gaussian one and then initialize the layer parameters manually, using their statistical estimations from the training data penultimate representation.
The calculation is performed as follows:
\begin{equation}
\mu_{c} = \frac{1}{|S_{c}|}\sum_{x\in S_{c}} f(x)
\end{equation}
\begin{equation}
\Sigma_{c} = \frac{1}{|S_{c}|}\sum_{x\in S_{c}} (\mu_{c} -f(x))(\mu_{c} -f(x))^T 
\end{equation}
\label{eq7}
where $S_{c}$ is the set which contains all of the samples from class $c$, $f(x)$ is the penultimate representation of $x$, and $\mu_{c}$ and $\Sigma_{c}$ are the layer parameters for class $c$.
After the layer's parameters are set, we fine-tune the model for 15 epochs using Equation (\ref{eq6}) as the optimization target. 
We stop the fine-tuning process when the validation loss starts to increase.

\subsection{Gaussian-Based Log Likelihood Ratio (LLR)}
The proposed DNN architecture outputs proper density values that correspond to the log likelihood of the classes' Gaussians in the penultimate representation.
Taking advantage of this property, we perform the $\mathcal{LLR}$ test on the Gaussians' log likelihood values. 
We define two scores, the predicted class ($p_{pred}$) log likelihood and the log likelihood average over the other classes ($p_{other}$):

\begin{equation}
 p_{pred}(x) = \max_{c\in \{1,\ldots, C\}} \log(\mathcal{N}(f(x); \mu_{c}, \Sigma_{c}))
\end{equation}

\begin{equation}
    p_{other}(x) = \frac{1}{C-1}\sum_{\substack{c'\neq c \\ c'\in \{1,\ldots,C\} }} \log(\mathcal{N}(f(x); \mu_{c'}, \Sigma_{c'}))
\end{equation}
We define the $\mathcal{LLR}$ score as:
\begin{equation}
    \mathcal{LLR}(x)  
    =p_{pred}(x)-p_{other}(x)
    \label{LLR}
\end{equation}
where $C$ is the number of classes, $x$ is the input sample, and $f(x)$ is the penultimate representation of $x$. 
We explain the $\mathcal{LLR}$ terms as follows: $p_{pred}$ is the log likelihood of a sample to be in the in-class distribution, and $p_{other}$ is the log likelihood of a sample to be out of the predicted class distribution.
The $\mathcal{LLR}$ (on its own) can serve as a basic and efficient score for OOD detection (Table~\ref{table:ablation}) that does not require proprietary tuning. 
In-distribution samples have higher $\mathcal{LLR}$ values, while OOD samples have lower $\mathcal{LLR}$ values, allowing them to be differentiated from one another.

\subsection{Artificial OOD Crafting}
\label{mimicing OOD}
One of the challenges that arises when creating an OOD detector is tuning it.
To address this challenge without additional data sources, we developed a novel method to generate artificial OOD samples that can be used to train an OOD detector.

We employ an adversarial crafting method to generate artificial OOD samples.
Given the training set samples (in-distribtion data), one can observe that some training samples have higher $\mathcal{LLR}$ values than others (Figure \ref{fig:LLR_dist}).
This implies that there are samples which are more likely to be associated with in-distribution samples than others. 
It is possible to use those values in order to estimate the boundaries of the in-distribution data. 
We estimate the boundary of the in-distribution data as the lower 95th percentile of the in-distribution data's $\mathcal{LLR}$ values; we refer to it as $Thres$.
We use an adversarial sample crafting method to alter the samples, such that their $\mathcal{LLR}$ values are lower than $Thres$, making those samples less likely to be associated with in-distribution data. 
We utilize the $LLR$ as the loss target when crafting the adversarial samples, iteratively altering the samples towards a lower $LLR$ value until $Thres$ is reached. 
The data used to perform this operation is a subset of the training data that was not used during training (a validation set); the algorithm is described in Algorithm~\ref{alg: gen}. 
\begin{algorithm}[t]
\small
\scalebox{0.9}{
\begin{minipage}{0.92\linewidth}
\label{Alg:Evolutionary_general}
    \begin{algorithmic}[0]
    \Procedure{Generate Sample}{x, $\epsilon$}
    \\
     \hspace*{\algorithmicindent} \textbf{Input}: $x$ - an input sample and $\epsilon$ - a step size
     \\
     \hspace*{\algorithmicindent} \textbf{Output}: $x_{art}$ an artificial OOD sample \\

    \State $x_{art}\leftarrow x$
    \For{$i \leq MaxIter$}\\
    \hspace*{\algorithmicindent} \#\textit{Add a perturbation}
        \State $x_{art}~\leftarrow~ x_{art}-\epsilon \nabla_{x} \mathcal{LLR} ( x_{art} ) $\\  \hspace*{\algorithmicindent} \#\textit{Clip to maintain valid image range}
        \State $x_{art}~\leftarrow~clip(x_{art})$
        \\ 
        \hspace*{\algorithmicindent}
         \#\textit{Stopping criteria}
        \If{ $\mathcal{LLR}(x_{art}) \leq Thres$ }
            \State Stop 
        \EndIf
    \EndFor
    \EndProcedure
    \caption{\label{alg: gen} }

    \end{algorithmic}
\end{minipage}
}
\end{algorithm}
$MaxIter$ was set at 10, because in most cases, the algorithm reaches its stopping criteria after 1-4 iterations.
The $\epsilon$ was not tuned based on OOD samples and was set to $0.01$ in all of our evaluations.
$Thres$ determines how large the perturbation will be for each sample, and it is calculated using only the in-distribution data. This eliminates the need to acquire  OOD samples prior to tuning our detector. A sensitivity analysis is presented in Table~\ref{table:sensitivity}, showing that $\epsilon$ and $Thres$ are insensitive to changes.

\subsection{Internal DNN Representation Feature Extraction}
\label{internal}
The internal DNN representations, which are the outputs of each layer's activation, are commonly used as input features for OOD detectors~\cite{mahalanobis, chen2020robust, sastry2019detecting}.
We were inspired by those studies and combined this approach with the maximum likelihood approach.
We extract the likelihood of the most likely class in each of the internal representations of the DNN and use it as features. 
In order to perform the likelihood calculation in each representation, we assume that the samples of each class are distributed as a Gaussian.
Under this assumption, we empirically estimate the parameters for the Gaussians (mean vector and diagonal covariance matrix) using the training data, as in Equation \eqref{eq3}.
To reduce the dimensionality, we apply a pooling function on the representations, instead of using the representations themselves. 
Unlike other OOD detection methods that only use average pooling over all representations, we use different pooling operations for different representations. 
Given a dataset, a sample $x$, and a DNN classifier of $C$ classes with $L+1$ hidden layers, for layers' representations $l \in \{1,\ldots, \frac{L}{2} \}$, we perform max pooling, and for $l \in \{\frac{L}{2}+1,\ldots, L\}$, we use average pooling. 
This modification was made because the clustering effect is more prominent in the deeper representations~\cite{katzir2019detecting}, and thus we assume that less noise reduction is required to extract useful information from them (i.e., average pooling can be used instead of max pooling).
Note that the feature extraction does not include the penultimate layer.
Our feature extraction algorithm does the following
\begin{equation}
    F(x, l) = \max_{c \in \{1, \ldots, C\}}  \log\big(\mathcal{N}(f^{[l]}(x); \mu^{[l]}_{c},\Sigma^{[l]}_{c})  \big)
    \label{eq8}
\end{equation}
where $F(x, l)$ is the feature extracted from layer $l$, i.e., the maximum log likelihood (Equation \ref{eq3}) estimated for a sample $x$ to belong to a certain class in the representation of layer $l$, using the diagonal covariance matrix $\Sigma^{[l]}_{c}$ and the mean vector $\mu^{[l]}_{c}$. $f^{[l]}(\cdot) \in\mathbb{R}^{d}$ is the output of the $l^{th}$ layer, $l\in \{1,\ldots, L\}$.

\subsection{OOD Detection}
\label{Detection}
The artificial OOD samples generated ~(\ref{alg: gen}) allow us to use a supervised learning classifier that differentiates between OOD samples and in-distribution samples. We train our classifier with artificial OOD samples labeled as OOD and a subset of the training data which was not used for the DNN training, which is labeled as in-distribution (the same samples that were used for the artificial OOD samples generation).
To achieve minimal inference time overhead, we employ a logistic regression classifier, implemented as an output neuron in the DNN, as our OOD detector. 
Given a sample $x$, the input to the detector is the internal representation features, as presented in Equation (\ref{eq8}), $ \{F(x, l)\}_{l=1}^{L}$, and the $\mathcal{LLR}$ presented in Equation (\ref{LLR}).
The score function of the logistic regression classifier is:
\begin{equation}
    score(x) = \sigma\big(\sum_{l=1}^{L} w_{l}\cdot F(x, l) + w_{L+1}\cdot \mathcal{LLR}(x) \big)
    \label{eq9}
\end{equation}
where $\sigma$ is the sigmoid function, and the $w_{l}$ are the weights of the logistic regression classifier.

%% file: Exp_setup.tex
\section{Experimental setup}
In this section, we provide details about our experimental process, evaluation measures, and datasets.

We evaluate FOOD's capabilities using three image datasets: CIFAR-10, CIFAR-100, and SVHN, which are considered the standard datasets for OOD detection. 
The datasets that the networks train on (in-distribution) are considered the positive class, and the OOD benchmarks, including LSUN, iSUN, and Tiny ImageNet, are considered the negative class (out-of-distribution).
We evaluate the proposed methods using the following commonly used metrics:

\begin{itemize}
\item TNR95: The true negative rate (TNR) when the true positive rate (TPR) is 95\%, which is a measure of how many OOD samples are detected correctly when we set the threshold so that the detector detects 5\% of the in-distribution as OOD. 

\item AUROC: The area under the receiver operating characteristic curve, which is a measure of performance for the detector over all possible detection thresholds.

\item Detection accuracy: The overall detection accuracy when adjusting the threshold for maximum detection between in-distribution samples and OOD samples.
\end{itemize}

We performed our evaluation using the PyTorch deep learning framework on a NVIDIA 2080 Ti GPU.
For the evaluation of FOOD's OOD detection performance, we trained five Resnet34 and five Resnet18\cite{he2016deep} models for each in-distribution dataset.
Each model was fine-tuned to have a Gaussian final layer, as presented in Equation~\ref{eq7}. 
As for optimization, we trained the models using SGD with a learning rate of $0.1$, momentum of $0.9$, and weight decay of $0.5e-4$. 
For the fine-tuning process, we used the RMSprop optimizer with a learning rate of $0.0001$ and selected the value for $\lambda$ ($\mathcal{R}_{ML}$ coefficient) using the validation set.
All of the other methods examined in our evaluation were implemented in PyTorch using the authors' code when possible.
For G.ODIN, MD, and MD*, we implemented the code ourselves. 
All methods (including ours) were based on the same 10 Resnet models. 
(r) postfix denotes that the mentioned dataset was used in its resized version.
Our code can be found on GitHub, along with the evaluation scripts of the other methods.\footnote{https://github.com/guyAmit/GLOD
}

%% file: Results.tex
\section{Results}

\subsection{Method Analysis}
\subsubsection{Ablation Study}

\input{tables/table2}

In Table \ref{table:ablation}, we evaluate the contribution of each of FOOD's stages to the overall method.
The stages appear in the \textit{Method} column in the table.
We present the simple MSP method \cite{he2016deep} as a baseline and the stages of our method are presented below it.
The first stage is $\mathcal{LLR}$ (stage 1), which uses a DNN whose final layer was replaced with a Gaussian layer.
To evaluate the OOD detection of this stage, we use the $\mathcal{LLR}$ value for OOD detection. 
The next stage is \textit{likelihoods sum + $\mathcal{LLR}$} (stage 2), which adds the sum of the likelihoods of the different representations, as presented in Section \ref{internal}.
At this stage, the method uses only average pooling on the internal representations for dimensionality reduction, before calculating the likelihood. 
This stage uses the sum of the likelihoods from each internal representation and the LLR score from the final layer as a score for OOD detection. 
Stage 3 introduces the logistic regression classifier (as presented in Section \ref{Detection}), where different weights are assigned to each likelihood value of the different representations, instead of simply summing them.
The only difference between this stage and the final version is the pooling method employed.
Finally, we present the full version of our method (stage 4).

We can see that in stage 1 the $\mathcal{LLR}$ obtains substantially better results compared to the MSP, implying that the $\mathcal{LLR}$ can be used as a preliminary OOD detection score.
In stage 2 the performance is further improved, suggesting that the use of internal representations benefits OOD detection. 
An even more substantial improvement appears in stage 3 where almost all of the tests show a 20\% performance gain for the TNR95 metric.
This reflects the critical improvement obtained by tuning the detector according to the artificial OOD samples (Section \ref{mimicing OOD}).
Finally, in stage 4, we see the addition of mixed pooling, which improves the performance on most datasets. 
Although the value of the TNR95 metric decreases on the SVHN dataset, the AUROC improves from stage 3 to stage 4 on this dataset. 
Overall, the mixed pooling is better when considering the OOD detection results in an aggregative manner over the different thresholds.
This indicates that our claim that less noise reduction is needed in deeper layers is correct.
\\

\subsubsection{Generating Artificial OOD Samples}
\begin{figure}[t]
    \centering
    \includegraphics[width=8cm]{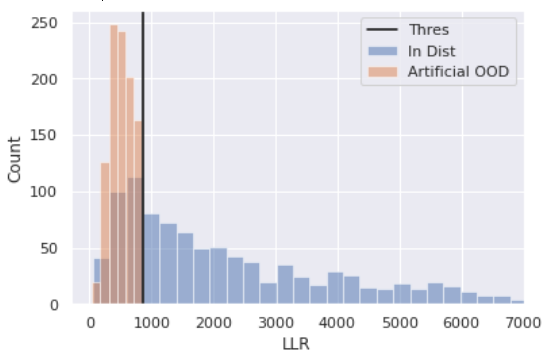}
    \caption{LLR values; CIFAR100 in-distribution training samples (blue) vs generated OOD samples (orange) crafted from the CIFAR100 model.}
    \label{fig:LLR_dist}
\end{figure}
To qualitatively demonstrate the effectiveness of the conversion of in-distribution samples to artificial OOD samples (Section \ref{mimicing OOD}), we measured the $\mathcal{LLR}$ values of each sample before and after converting it into its artificial OOD version.
Figure \ref{fig:LLR_dist} presents how the samples behave in terms of the $\mathcal{LLR}$ value.
The $\mathcal{LLR}$ values of the vanilla samples from the CIFAR100 validation dataset appear in blue, and the $\mathcal{LLR}$ values of converted samples created from the same samples appear in orange.
As shown in the figure, after conversion, almost all of the samples obtain an $\mathcal{LLR}$ value lower than the 95th percentile bound (as presented in Section \ref{mimicing OOD}).
\\

\subsubsection{Sensitivity Analysis}
\input{tables/table4}

In this subsection, we show that FOOD's artificial OOD generation method is resilient to to the choice of its hyperparameters, namely the step size $\epsilon$ and the threshold $Thres$.
Table~\ref{table:sensitivity} presents the detection results for a range of hyperparameters choices.
As shown in the table, the detection measures are not significantly affected by different hyperparameters choices.
This shows that the $\epsilon$ and $Thres$ hyperparamters are not sensitive to changes within a reasonable range.

\subsection{Inference Time Comparison}
\begin{figure}[b]
    \centering
    \includegraphics[width=9cm]{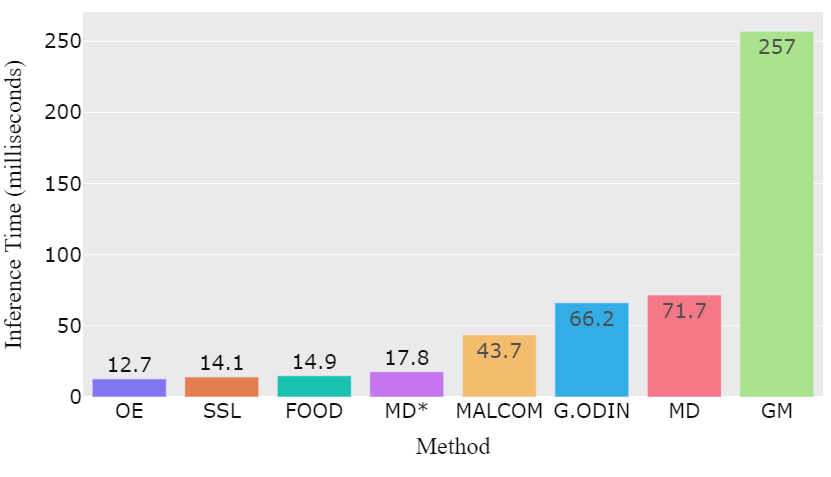}
    \caption{Inference time comparison (times measured for the DNN inference and OOD detection combined); all methods' inference times were measured on the CIFAR100 dataset with five different Resnet34 models and averaged.}
    \label{fig:Speed_comparison}
\end{figure}
Figure \ref{fig:Speed_comparison}, which provides a comparison of the inference times of state-of-the-art OOD detection methods, shows that there is a large variation between the methods. 
Unsurprisingly, methods that rely on MSP had the fastest inference time, as the only calculation required for them is the feedforward operation. 
FOOD's inference time is similar to those methods and is also the fastest of all of the methods that do not require OOD samples for training. 
MD* is a little slower; this can be explained by the use of full covariance matrices in this method, as opposed to \MethodName which uses diagonal covariance matrices.
We observe that methods which require adversarial preprocessing (e.g., G.ODIN, MD) are more than four times slower than FOOD, as they require backpropagation in order to edit the sample before inference. 
GM requires complicated postprocessing, causing its run to take the longest of the methods examined.

\begin{figure}[b]
    \centering
    \includegraphics[width=8cm]{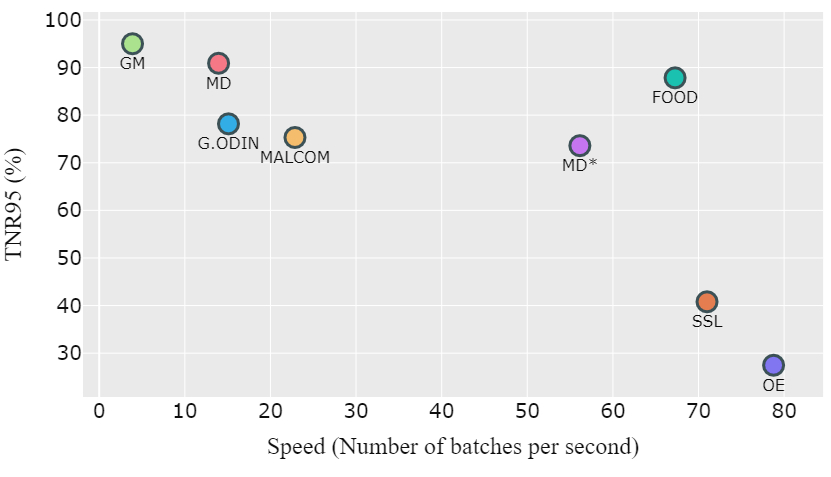}
    \caption{${Speed}$ vs $TNR95$ comparison (times measured for the DNN inference and the OOD detection combined); all methods' inference times and OOD detection performance were measured on the CIFAR100 dataset with five different ResNet34 models and averaged.}
    \label{fig:SpeedVsTNR}
\end{figure}

Figure~\ref{fig:SpeedVsTNR} shows the trade-off between OOD detection performance and inference time for different methods, evaluating the TNR95 metric against the speed. 
The figure shows that methods that are both fast and accurate tend to appear closer to the top right corner, whereas methods with suboptimal detection performance or a long inference time appear respectively more to the left and closer to the $Speed$ axis. 
As seen in the plot, FOOD's inference time is faster than almost all the other methods, while maintaining high OOD detection capabilities. 
Methods that obtain higher TNR95 scores run significantly slower, and in the case of MD, use OOD samples for tuning. 
In addition, it is easy to see that methods which have a faster inference time than \MethodName (i.e., OE, and SSL) have significantly lower OOD detection performance.

\subsection{OOD Detection Performance Comparison}
%
\input{tables/table3}
Table \ref{table:main} presents a comprehensive OOD detection performance comparison of methods that meet the \textit{Fast Inference Time} requirement (Section \ref{restrictions}). 
As seen in the table, \MethodName consistently obtains better results on the TNR95 metric, outperforming the other methods on all datasets on this metric.
The smallest gap in performance was seen on the SVHN dataset, although the other metrics (AUROC and detection accuracy) indicate that there is a significant difference between the methods on that dataset as well.
The most substantial performance gap between the methods examined is seen on the CIFAR100 dataset. DNN classification performance on CIFAR100 tend to be much lower than CIFAR10 and SVHN, this can affect the OOD detector's performance. 
We observe that the results obtained by FOOD and MALCOM (the method achieving the second best OOD detection results) for the TNR95 metric on the CIFAR100 dataset are respectively $88.9$ (on average) and $75.2$. 
Resnet18-based models showed similar trends; on the CIFAR100 dataset, the results for the TNR95 metric are $91$ (on average) for FOOD and $76.9$ for MALCOM, which again has the second best OOD detection results.
Due to space limitations, we cannot present the full evaluation of the methods based on Resnet18.

%% file: tables/table2.tex
\begin{table*}[h]
\begin{center}
\caption{Ablation Study - Resnet34 CIFAR100 results of the different stages of the method}
\begin{tabular}{cccc}
\hline
\multirow{2}{*}{Method}                                               & \textbf{TNR95}    & \textbf{AUROC} & \textbf{Detection Accuracy}  \\ \cline{2-4} 
                                                                                                                                                                                                   & \multicolumn{3}{c}{SVHN / LSUN(r) / Tiny-Imagenet(r) / Isun(r)}                                                                                                                                                                                                                                                                                                                                                                                                                                                                                           \\ \hline
\multicolumn{1}{l}{\begin{tabular}[c]{@{}l@{}}Baseline - MSP\\ Stage 1 - $\mathcal{LLR}$\\ Stage 2 - Likelihood sum + $\mathcal{LLR}$\\ Stage 3 - FOOD - Avg Pooling\\ Stage 4 - FOOD - Mixed Pooling (final)\end{tabular}} & \begin{tabular}[c]{@{}c@{}}19.5 / 25.8 / 27.5 / 26.0\\ 56.7 / 34.7 / 28.6 / 29.4\\ 66.3 / 71.1 / 61.5 / 64.4\\ 86.8 / 90.5 / 84.8 / 87.8\\ 74.7 / 96.0 / 91.6 / 93.4\end{tabular} & \begin{tabular}[c]{@{}c@{}}79.3 / 81.1 / 80.6 / 80.4\\ 92.4 / 85.9 / 82.5 / 83.9\\ 93.6 / 92.8 / 89.7 / 90.0\\ 93.7 / 97.7 / 96.2 / 96.9\\ 95.5 / 98.9 / 97.8 / 98.3\end{tabular} & \begin{tabular}[c]{@{}c@{}}73.2 / 73.8 / 73.1 / 73.8\\ 85.3 / 78.7 / 75.8 / 77.1\\ 85.9 / 84.7 / 80.9 / 82.3\\ 91.9 / 92.8 / 90.1 / 91.6\\ 88.5 / 96.2 / 93.6 / 94.6\end{tabular} \\ \hline
\end{tabular}
\label{table:ablation}
\end{center}
\end{table*}

%% file: tables/table4.tex
\begin{table}[t]
\centering
\caption{Sensitivity Analysis - Resnet34 CIFAR100 models. OOD detection results are averaged over four OOD datasets.}
\begin{tabular}{cc}
\hline
TNR95                        & AUROC                     \\ \hline
\multicolumn{2}{c}{$\epsilon$ :  0.04/ 0.02 / 0.01 / 0.005 / 0.0025} \\ \hline
87.3 / 85.8 / 88.9 /  88.2 /  89.2          & 97.2 / 97.0 / 97.6 / 97.5 / 97.6         \\ \hline
\end{tabular}
\label{table:sensitivity}
\end{table}

\begin{table}[t]
\centering
\begin{tabular}{cc}
\hline
TNR95                        & AUROC                     \\ \hline
\multicolumn{2}{c}{$thres :  90\%/ 92.5\% / 95\% / 97.5\% / 99\%$} \\ \hline
89.2 / 89.0 / 88.9 /  88.7 /  86.6          & 97.7 / 97.7 / 97.6 / 97.5 / 97.0         \\ \hline
\end{tabular}
\end{table}

%% file: tables/table3.tex
\begin{table*}[h]
\centering
\caption{OOD detection performance comparison}
\begin{tabular}{ccccl}
\hline
\multirow{2}{*}{$D_{in}$} & \multirow{2}{*}{$D_{out}$}                                                              & \textbf{TNR95}                                                                                                                                                                       & \textbf{AUROC}                                                                                                                                                                                          & \multicolumn{1}{c}{\textbf{Detection Accuracy}}                                                                                                                                                       \\ \cline{3-5} 
                          &                                                                                         & \multicolumn{3}{c}{OE / SSL / MD* /  Malcom / FOOD}                                                                                                                                                                                                                                                                                                                                                                                                                                                                                                                                                  \\ \hline
CIFAR-100                 & \begin{tabular}[c]{@{}c@{}}SVHN\\ LSUN(r)\\ Tiny-Imagenet(r)\\ Isun(r)\end{tabular}     & \begin{tabular}[c]{@{}c@{}}34.0 / 58.6 / 56.0 / 63.7 / \textbf{74.7} \\  31.7 / 41.2 / 80.4 / 80.0 / \textbf{96.0}\\  19.7 / 26.6 / 79.2 / 79.3 / \textbf{91.6}\\  24.3 / 36.8 / 78.7 / 78.1 / \textbf{93.4}\end{tabular}    & \begin{tabular}[c]{@{}c@{}}85.5 / 93.0 / 91.5 / 92.8 / \textbf{95.5}  \\ 70.9 / 87.9 / 96.5 / 96.2 / \textbf{98.9}\\ 79.7 / 82.3 / 96.1/ 96.1 / \textbf{97.8}\\ 67.1 / 86.6 / 96.0 / 95.8 / \textbf{98.3}\end{tabular}                       & \begin{tabular}[c]{@{}l@{}}78.3 / 85.9 / 83.9 / 85.0 / \textbf{88.5}\\ 73.0 / 81.4 / 90.1 / 89.4 / \textbf{96.2}\\ 67.1 / 76.3 / 89.3 / 89.0 / \textbf{93.6}\\ 69.6 / 80.0 / 89.3 / 88.8 / \textbf{94.6}\end{tabular}                      \\ \hline
CIFAR-10                  & \begin{tabular}[c]{@{}c@{}}SVHN\\ LSUN(r)\\ Tiny-Imagenet(r)\\ Isun(r)\end{tabular}     & \begin{tabular}[c]{@{}c@{}}96.6 / 97.5 / 69.0 / 73.7 / \textbf{98.1}\\  93.3 / 95.3 / 92.1 / 90.4 / \textbf{99.5}\\  81.7 / 87.9 / 89.0 / 87.3 / \textbf{98.8}\\  93.1 / 94.6 / 90.6 / 89.3 / \textbf{99.3}\end{tabular}     & \begin{tabular}[c]{@{}c@{}}99.0 / 99.3 / 95.6 / 96.1 / \textbf{99.5}\\ 98.5 / 98.8 / 98.4 / 98.4 / \textbf{99.8}\\ 96.6 / 97.6 / 98.1 / 98.0 / \textbf{99.7}\\ 98.4 / 98.7 / 98.2 / 98.2 / \textbf{99.8}\end{tabular}                        & \multicolumn{1}{c}{\begin{tabular}[c]{@{}c@{}}96.0 / 96.5 / 91.0 / 91.3 / \textbf{97.2}\\ 94.8 / 95.5 / 94.3 / 94.0 / \textbf{98.5}\\ 91.2 / 92.9 / 93.2 / 92.9 / \textbf{97.7}\\ 94.5 / 95.2 / 93.6 / 93.5 / \textbf{98.2}\end{tabular}} \\ \hline
SVHN                      & \begin{tabular}[c]{@{}c@{}}CIFAR-10\\ LSUN(r)\\ Tiny-Imagenet(r)\\ Isun(r)\end{tabular} & \begin{tabular}[c]{@{}c@{}}94.2 / 93.6 / 87.7 / 93.2 / \textbf{97.9}\\  93.1 / 93.5 / 98.0 / \textbf{99.1} / \textbf{99.1}\\  93.8 / 93.9 / 96.7 / 98.4 / \textbf{99.3}\\  93.7 / 94.0 / 97.6 / 99.0 / \textbf{99.2}\end{tabular} & \multicolumn{1}{l}{\begin{tabular}[c]{@{}l@{}}98.8 / 98.9 / 97.4 / 98.2 / \textbf{99.4}\\  98.5 / 98.8 / 99.0 / 99.4 / \textbf{99.8}\\  98.8 / 98.9 / 98.9 / 99.3 / \textbf{99.8}\\  98.7 / 98.9 / 99.0 / 99.5 / \textbf{99.8}\end{tabular}} & \begin{tabular}[c]{@{}l@{}}97.6 / 97.7 / 93.1 / 94.6 / \textbf{96.8}\\ 97.4 / 97.7 / 96.6 / 97.6 / \textbf{98.6}\\ 97.7 / 97.8 / 95.8 / 97.0 / \textbf{98.5}\\ 94.0 / 97.8 / 96.3 / 97.5 / \textbf{98.5}\end{tabular}                     \\ \hline
\end{tabular}
\label{table:main}
\end{table*}

%% file: Conclusions.tex
\section{Conclusion}
In this paper, we addressed the OOD detection problem from a real-world application perspective.
We introduced FOOD, an OOD detector method with state-of-the-art OOD detection performance and fast inference time that does not require additional training data. 
To the best of our knowledge, this is the first paper to compare the inference time of state-of-the-art OOD detection methods.
Our OOD detection performance evaluations demonstrate that \MethodName outperforms the other fast state-of-the-art OOD detectors.
This makes \MethodName more suitable for real-world applications than other OOD detection methods.
Our method for generating OOD samples captures the perceived boundaries of the training data from the DNN perspective. 
Thus, beyond generating artificial OOD data, it can be used for data augmentation, enabling samples to be generated in a deliberate manner.
Moreover, we hypothesize that other OOD detection methods can employ our data generation process to eliminate the need to acquire OOD data.
In future work, we plan to evaluate \MethodName in conjunction with unsupervised learning methods to enhance the representations learned by the model.